%% file: IPIQA.tex
\documentclass{article}
\usepackage{spconf,amsmath,epsfig}
\usepackage{amsfonts}
\usepackage{algorithmic}
\usepackage{algorithm}
\usepackage{array}
\usepackage[caption=false,font=normalsize,labelfont=sf,textfont=sf]{subfig}
\usepackage{hyperref}

\usepackage{textcomp}
\usepackage{stfloats}
\usepackage{url}
\usepackage{verbatim}
\usepackage{graphicx}
\usepackage{cite}

\usepackage{multirow}
\usepackage{booktabs}
\usepackage{makecell}

\usepackage{setspace}

\let\OLDthebibliography\thebibliography
\renewcommand\thebibliography[1]{
  \OLDthebibliography{#1}
  \setlength{\parskip}{0pt}
  \setlength{\itemsep}{0pt plus 0.3ex}
}

\begin{document}\sloppy

% Example definitions.
% --------------------
\def\x{{\mathbf x}}
\def\L{{\cal L}}

% Title.
% ------
% \title{IP-IQA: A Multimodal Baseline for AI-Generated Image Quality Assessment via Corresponding Image and Prompt}
% \title{IP-IQA: A Multimodal Baseline for AI-Generated Image Quality Assessment via Integration Textual Prompt}

\title{Bringing Textual Prompt to AI-Generated Image Quality Assessment}
%
% Single address.
% ---------------
% \name{Anonymous ICME submission}
% %Address and e-mail should NOT be added in the submission paper. They should be present only in the camera ready paper. 
% \address{PaperID: 2098}

\name{Bowen Qu$^{\ast}$, Haohui Li$^{\ast}$, Wei Gao$^{\dagger}$}

\address{School of Electronic and Computer Engineering, Peking University, China \\ $^{\ast}$\{bowenqu, lihaohui\}@stu.pku.edu.cn \ \ \ \  $^{\ddagger}$gaowei262@pku.edu.cn}

\maketitle

\setstretch{0.92}
\begin{abstract}
AI-Generated Images (AGIs) have inherent multimodal nature. Unlike traditional image quality assessment (IQA) on natural scenarios, AGIs quality assessment (AGIQA) takes the correspondence of image and its textual prompt into consideration. This is coupled in the ground truth score, which confuses the unimodal IQA methods. To solve this problem, we introduce IP-IQA (AG\textbf{I}s \textbf{Q}uality \textbf{A}ssessment via \textbf{I}mage and \textbf{P}rompt), a multimodal framework for AGIQA via corresponding image and prompt incorporation. Specifically, we propose a novel incremental pretraining task named Image2Prompt for better understanding of AGIs and their corresponding textual prompts. An effective and efficient image-prompt fusion module, along with a novel special \textit{[QA]} token, are also applied. Both are plug-and-play and beneficial for the cooperation of image and its corresponding prompt. Experiments demonstrate  that our IP-IQA achieves the state-of-the-art on AGIQA-1k and AGIQA-3k datasets. Code will be available at \href{https://github.com/Coobiw/IP-IQA}{https://github.com/Coobiw/IP-IQA}.
\end{abstract}

\renewcommand{\thefootnote}{}
\footnotetext{$^{\ast}$Equal Contribution. $^{\dagger}$Corresponding author.\ This work was supported by Natural Science Foundation of China (62271013, 62031013), Guangdong Province Pearl River Talent Program High-Caliber Personnel - Elite Youth Talent (2021QN020708), Shenzhen Fundamental Research Program (GXWD20201231165807007-20200806163656003), Shenzhen Science and Technology Program (JCYJ20230807120808017), and Sponsored by CAAI-MindSpore Open Fund, developed on OpenI Community (CAAIXSJLJJ-2023-MindSpore07).}

\section{Introduction}
\label{sec:intro}
\input{intro_v2}

\input{related_work}

\input{method}

\input{experiment}

\input{conclusion}

\small
\bibliographystyle{IEEEbib}
\bibliography{IEEEabrv, AGIQA_REF}

\end{document}

%% file: intro_v2.tex
% The content generated with the assistance of AI technology is referred to Artificial Intelligence Generated Content (AIGC), including images, audios, texts, and videos. Recently, numerous AI-Generated Image (AGI) models based on different technical routes have been developed, mainly branched into GAN\cite{goodfellow2014generative}, auto regressive-based models\cite{ding2021cogview} and diffusion-based models\cite{rombach2022high}, providing different choices for generating images. As an emerging new image type consuming by end-users, the quality of AGIs need assessing to provide a good visual experience. Thus, it is significant to conduct objective image quality assessment (IQA) study for AGIs.

Artificial Intelligence Generated Content (AIGC), including images, texts and videos generation, is booming. Numerous AI-Generated Image (AGI) models based on different technical routes have been developed, mainly branched into GAN\cite{goodfellow2014generative}, auto regressive-based models\cite{ding2021cogview} and diffusion-based models\cite{rombach2022high}. As an emerging new image type, AGIs need comprehensive quality assessment (QA) for better visual experience.

% There are many advanced IQA methods proposed for evaluating natural scene images (NSIs). Wang \textit{et al.} \cite{wang2023exploring} explores the potential of CLIP \cite{radford2021learning} for assessing both the quality (look) and abstract perception (feel) of images in a zero-shot manner. Su \textit{et al.}\cite{su2020blindly} introduced a self-adaptive hyper network named HyperNet. However, these models only take images as input, which is not sufficient to evaluate the quality of AGIs. The generation models' understanding of the input text prompts relies on the language model, which makes the correspondence between generated image and text vary. And it leads to certain distortions about image-text discrepancy. Therefore, the absence of textual modality in the assessment of quality stage can lead to failure in evaluating AGI, even though these methods take into account the human visual system (HVS).

% Wang \textit{et al.} \cite{wang2023exploring} explores the potential of CLIP \cite{radford2021learning} for assessing both quality and abstract perception, using hard template texts.

There are many advanced IQA methods proposed for evaluating natural scene images (NSIs). Su \textit{et al.}\cite{su2020blindly} introduced HyperNet which is self-adaptive. Zhang \textit{et al.} \cite{zhang_blind_2020} introduces a deep bilinear convolutional neural network named DBCNN. Multimodal models are also explored to enhance the assessing performance. However, AGIs are inherently multimodal entities, each accompanied by a corresponding textual prompt. These models only take images as input actually, which is not sufficient to evaluate the quality of AGIs with the absence of whole textual prompts understanding.

We perform a toy experiment to use the IQA method (i.e., ResNet50) to evaluate the image quality of AGIs. The experiment results are shown in Fig. 1. The ground truth score includes considerations of image quality and the image-prompt correspondence. As seen, although these images exhibit high visual quality, the IQA method fall short in terms of image-text correspondence. That is the reason that ResNet50 predicts significantly higher scores than the actual ground truth. Thus, we should explore how to integrate textual prompts into the AGIQA framework to achieve a comprehensive assessment.

\begin{figure}[!t]
\centering
\includegraphics[width=1\linewidth]{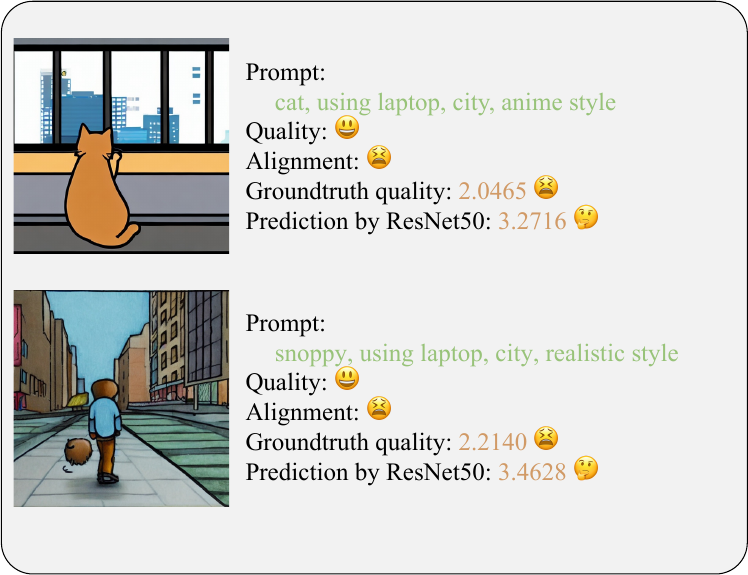}
% \caption{Analysis of bad and good cases by ResNet50 on the AGIQA-1k dataset. These demonstrate the unimodal IQA model's tendency to assess image quality without the capability to analyze the correspondence between image and text, as evidenced by the absence of prompt input. This limitation leads to a disconnect in evaluating the ground truth scores coupled with image-text correspondence.}
\caption{Quality assessment results generated by ResNet50 on the AGIQA-1k dataset. As seen, ResNet50 tends to assess image quality without analyzing the correspondence between image and text prompt, generating unsatisfactory assessment scores.}
\label{cases}
\end{figure}

% To address the problems, a multimodal AGIQA model is proposed, which can hanle both image and its corresponding textual prompt, named IP-IQA (AG\textbf{I}s \textbf{Q}uality \textbf{A}ssessment via \textbf{I}mage and \textbf{P}rompt). The incorporation of textual prompt aligns well with the inherently multimodal nature of AGIs. 

Inspired by the success of CLIP model~\cite{radford2021learning} for multi-modal learning, we introduce a CLIP-based dual-stream framework named IP-IQA (AG\textbf{I}s \textbf{Q}uality \textbf{A}ssessment via \textbf{I}mage and \textbf{P}rompt) for processing corresponding AGIs and textual prompts simultaneously. Specifically, we initialize the image and text encoder by the original pretrained CLIP weights. However, CLIP is trained on a large-scale web image-text dataset, which exhibits significant divergence from the distribution of AGIs. Thus, we construct an Image2Prompt pretraining task to pretrain the image encoder  incrementally on a subset of AI-Generated Images database DiffusionDB\cite{wang2022diffusiondb}. Besides, for better image-text interaction,  we propose an image-prompt fusion module and the special \textit{[QA]} token, and insert them to the pretrained model. Extensive experiments are performed on AGIQA-1k and AGIQA-3k datasets. The results demonstrate the effectiveness of our method. To summarize, our contributions are three-fold:
\textbf{1)} We propose an incremental pretraining task termed Image2Prompt, which is significantly beneficial for the multimodal quality assessment model to understand the AGIs and their corresponding textual prompts; \textbf{2)} We propose an effective and efficient image-prompt fusion module, as well as a special design quality assessment token, for learning comprehensive representations for AGIQA; \textbf{3)} IP-IQA achieves the state-of-the-art performance on both AGIQA-1k\cite{zhang2023perceptual} and AGIQA-3k\cite{li2023agiqa}. To best of our knowledge, we are the first work to take both image and text into consideration in AGIQA community.

% The key contributions of our work can be summarized as following: \textbf{1)} IP-IQA is the first work to take both image and the whole textual prompt into consideration for comprehensive assessment, to our best knowledge in AGIQA community. \textbf{2)} IP-IQA introduces an incremental pretraining task, named Image2Prompt, which is significantly beneficial for the initial CLIP model to understand the AGIs and their corresponding prompts. This period is performed on a subset of DiffusionDB\cite{wang2022diffusiondb} database. We use CLIP to filter the low-quality image-text pairs, keeping 560K for our training. \textbf{3)} IP-IQA also proposes an effective and efficient image-prompt fusion module, as well as a special \textit{[QA]} token, which are both plug-and-play. \textbf{4)} IP-IQA achieves the state-of-the-art performance on both AGIQA-1k\cite{zhang2023perceptual} and AGIQA-3k\cite{li2023agiqa}.

%% file: related_work.tex
\section{Related Work}
\label{sec:related work}

\subsection{AGI Quality Assessment Methods}
To evaluate the quality of AGIs, several quantitative evaluation metrics have been proposed, mainly focusing on assessing perceptual quality and the T2I correspondence. In terms of the perceptual quality, Inception Score (IS)\cite{salimans2016improved} and Fréchet Inception Distance (FID)\cite{hessel2021clipscore} are employed for the performance measurement of the generation model. IS evaluates the sharpness and diversity of the generated images by analyzing the class probabilities obtained using Inception-V3. And FID measures the difference in feature distribution between a set of generated images and a set of real-world images. In the perspective of the T2I correspondence, CLIPScore\cite{hessel2021clipscore} is mainly used to evaluate the quality of images generated by text-to-image models, leveraging the capabilities of CLIP\cite{radford2021learning}. It assesses the alignment between generated images and their corresponding textual descriptions by calculates the similarity between the embeddings of the text and the embeddings of the generated image. However, these methods are heavily reliant on specific datasets or pre-trained models, especially IS. Besides, these metrics can be sensitive to their calculation parameters, such as batch size, leading to variability in their reliability and effectiveness. Crucially, they do not incorporate considerations of the human visual system, which means they may not align with human perception in assessing image quality or relevance.

\subsection{Deep-based Image Quality Assessment Methods}
Kang \textit{et al.} \cite{kang2014convolutional} pioneers the use of deep convolutional neural networks for no-reference image quality assessment (NR-IQA). Their method leverages a CNN to directly learn image quality representations from raw image patches, without relying on hand-crafted features or a reference image. Zhang \textit{et al.} \cite{zhang_blind_2020} introduces a deep bilinear convolutional neural network for blind image quality assessment (BIQA), uniquely combining two CNN streams to separately address synthetic and authentic image distortions. Su \textit{et al.}\cite{su2020blindly} introduced a self-adaptive hyper network named HyperNet. This method innovatively assesses the quality of authentically distorted images through a three-stage process: content understanding, perception rule learning, and quality prediction. There are many other representative NR-IQA methods including: CLIP-based vision-language correspondence \cite{wang2023exploring, zhang2023liqe} and loss function for 
fast convergence\cite{lilinearityiqa}. However, the absence of textual modality in the assessment of quality stage can lead to failure in evaluating AGI, even though these methods take into account the human visual system. 

%% file: method.tex
\section{Methodology}
\label{sec:method}

% \begin{figure*}[!t]
%     \centering
%     \subfloat[]{
%         \includegraphics[width=2.3in]{figs/i2p_framework.png}
%         \label{fig:img2prompt}
%     }
%     \hfil
%     \subfloat[]{
%         \includegraphics[width=2.3in]{figs/framework.png}
%         \label{fig:framework}
%     }
%     \hfil
%     \subfloat[]{
%         \includegraphics[width=2.2in]{figs/attn_pooling.pdf}
%         \label{fig:attention_pooling}
%     }
%     \caption{ Detailed overview of the IP-IQA framework. (a) is our Image2Prompt incremental pretraining framework. (b) illustrates the IP-IQA framework featuring a modular image-prompt fusion component with the novel \textit{[QA]} special token designed for quality assessment. (c) depicts the distinction between the Attention Pooling module and the Cross-Modality Attention Pooling module, highlighting the variation in the global query token.}
%     \label{fig_arch}
% \end{figure*}

\begin{figure*}[!t]
\centering
\includegraphics[width=1\linewidth]{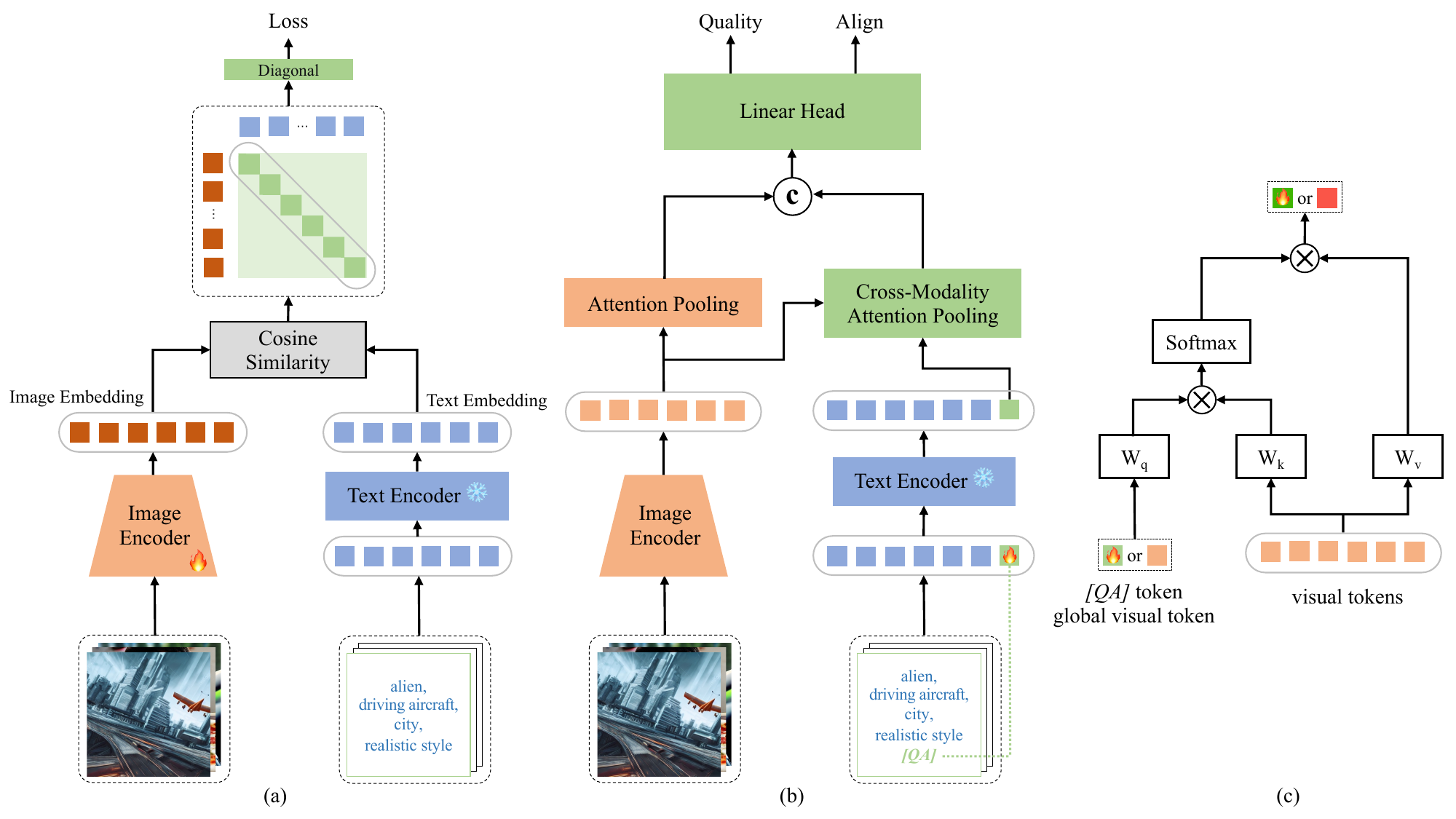}
\caption{Detailed overview of the IP-IQA framework. (a) presents the Image2Prompt incremental pretraining framework. (b) illustrates the IP-IQA framework featuring a modular image-prompt fusion component with the trainable \textit{[QA]} special token designed for quality assessment. (c) shows the workflows of Attention Pooling module and the Cross-Modality Attention Pooling module, highlighting the variation in the global query token. The global visual token is computed by spatially global average pooling (GAP) operation.}
\label{fig_arch}
\end{figure*}

% \subsection{Overview}
% Our IP-IQA can be structured into three core components as following:

% \textbf{Integration of Textual Prompts:} This section briefly explains how to introduce multi-modality for AGIQA. We use a dual-stream architecture to handle image and its corresponding textual prompt simultaneously.

% \textbf{Image2Prompt Incremental Pre-training:} We introduce \textit{Image2Prompt}, an incremental pretraining method on initial CLIP model, tailored for the vision-language multimodal nature of AGIs. This method bridges the AGI-style visual and textual modalities, enhancing the model's multimodal understanding which is essential for AGIQA.

% \textbf{Image-Prompt Fusion Module \& \textit{[QA]} Token:} We propose an effective and efficient cross-attention based module for image-text fusion. This module, along with the special \textit{[QA]} token in the prompt textual domain, focus our model on quality-relevant aspects of both image and its corresponding prompt. It's a good way to build images and prompts with a good integration and interaction on AGIQA tasks.

\subsection{Overview}
The overview of the proposed IP-IQA is illustrated in Fig. 2. It is a dual-stream architecture to simultaneously process the image and its corresponding textual prompt. Specifically, IP-IQA is constructed based on the CLIP model. To enhance the model's multimodal understanding, we introduce the Image2Prompt strategy, an incremental pretraining method on the initial CLIP model tailored for the vision-language multimodal nature of AGIs. This strategy bridges the AGI-style visual and textual modalities, which is essential for AGIQA. Besides, to build images and prompts with good integration and interaction for AGIQA, we propose a cross-attention-based image-prompt fusion module along with the specially designed \textit{[QA]} token in the textual prompt domain, guiding the model on quality-relevant aspects of both image and its corresponding textual prompt. Next, we will analyse the necessity of integration of textual prompts.

\subsection{Integration of Textual Prompts}
AGIs have inherent multimodal natures from birth. This intrinstic characteristic forms the foundation of our approach towards AGIQA. Traditional AGIQA methods predominantly follow the trajectory of IQA, which only operates in a unimodal context. 

The ground truth score of AGIQA-1k dataset\cite{zhang2023perceptual} takes image quality, aesthetics and image-text correspondence into consideration. However, IQA models tend to focus on the visual quality and aesthetics, failing to appropriately assess the correspondence between the image and its textual prompt. The case of the two images in Fig. 1 illustrate this. It is of great necessity of integrating textual prompts into the AGIQA framework to achieve a comprehensive assessment.

As shown in Fig. 2(a) and Fig. 2(b), we produce a CLIP-like\cite{radford2021learning} dual-stream architecture with separate encoders to process the image and textual prompt inputs respectively. Both are initialized by CLIP model. We can separately get image and prompt embedding. Given an image $I$ and its corresponding textual prompt $T$, let $f_{\theta_{\text{img}}}(I)$ represent the embedding produced by the image encoder with parameters $\theta_{\text{img}}$, and let $h_{\theta_{\text{txt}}}(T)$ represent the embedding produced by the text encoder with parameters $\theta_{\text{txt}}$. Additionally, we utilize the embedding of \textit{[eot]} (end the text) token to represent the entire prompt sentence.

\subsection{Image-to-Prompt Incremental Pre-training}

To meet with the multimodal nature of AGIs, we propose an incremental pretraining method named Image2Prompt on CLIP-initialized model. CLIP\cite{radford2021learning} is trained on 400M image-text pairs from the Internet, which have domain gaps with AGIs. This method is specifically tailored to bridge the gap between AGI-style visual and textual modalities, thereby enhancing the model's understanding of the complex interplay between an AGI and its corresponding prompt.

We freeze the CLIP text encoder to get stable prompt embedding space and train the image encoder by cosine similarity loss between pair of corresponding embeddings, given by:
\begin{equation}
\label{Image2Prompt Loss}
L_{cos}(I, T) = 1 - \frac{f_{\theta_{\text{img}}}(I) \cdot h_{\theta_{\text{txt}}}(T)}{\|f_{\theta_{\text{img}}}(I)\| \|h_{\theta_{\text{txt}}}(T)\|}
\end{equation}

This loss function shown in Eq. \ref{Image2Prompt Loss} ensures that the embeddings of AGI and its corresponding prompt are aligned in similar space, promoting image-prompt matching.

% \textit{Image2Prompt} incremental pre-training requires a substantial corpus of AGI-prompt pairs. Hence, we utilize the subset of DiffusionDB\cite{wang2022diffusiondb}, which contains 2 million image-prompt pairs. Then, we calculate the cosine similarity between image and its corresponding prompt by CLIP, filtering the sample whose score is lower than a predefined threshold. This strategy ensures that our incremental pretrain procedure is informed by high-quality, well-matched image-prompt pairs, thereby enhancing the efficacy of the model in capturing the interplay between the two modalities.

% \subsection{Image-Prompt Fusion Module \& \textit{[QA]} Token}

\subsection{Image-Prompt Fusion Module}
In order to combine the image and its corresponding prompt effectively and efficiently, we propose a plug-and-play image-prompt fusion module. Given an $(image, prompt)$ pair represented by $(I, T)$. We use $\theta^{\prime}_{\text{img}}$, $\theta_{\text{txt}}$ to represent the image encoder continually pretrained by our Image2Prompt and the CLIP text encoder correspondingly. Thus, the visual embedding $Embed_v$ and textual prompt embedding $Embed_t$ can be computed by:
\begin{equation}
\label{eq:2}
\begin{aligned}
Embed_v &= f_{\theta^{\prime}_{\text{img}}}(I) \\
Embed_t &= h_{\theta_{\text{txt}}}(T, \textit{[QA]})
\end{aligned}
\end{equation}
\noindent where \textit{[QA]} refers to a special token designed for AGIQA. Originally, the token in this position should be \textit{[eot]} token, which means "end of text". We replace the \textit{[eot]} by the \textit{[QA]} special token to pay more attention to vocabulary related to image quality during the extraction of prompt embedding. Additionally, all parts the text encoder are frozen, except our \textit{[QA]} token. The global feature of visual embedding is computed by spatially global average pooling, represented by $GAP$. As for its corresponding prompt, we use the embedding of \textit{[QA]} token, which is at the end of sentence, for global encoding.
\begin{equation}
\label{eq:3}
\begin{aligned}
G_v &= GAP(Embed_v) \\
G_t &= Embed_t[:,-1,:] 
\end{aligned}
\end{equation}

Then, we serve these two as queries for their corresponding attention pooling layers to get pure visual features and cross-modality fused features. The details are shown as following:

\begin{equation}
\label{eq:4}
\text{Attention}(Q, K, V) = \text{softmax}\left(\frac{QK^T}{\sqrt{d_k}}\right)V
\end{equation}

\noindent where
\begin{equation}
\label{eq:5}
\begin{aligned}
Q &= W_q \cdot \{G_v,G_t\}  \\
K &= W_k \cdot (Embed_v + Embed_{pos}) \\
V &= W_v \cdot (Embed_v + Embed_{pos})
\end{aligned}    
\end{equation}
Here, $d_k$, $Embed_{pos}$ refers to the number of channels and learnable positional embedding of image patches. The $W_q$, $W_k$ and $W_v$ refer to the projection matrix of query, key and value correspondingly. It should be noted that the query is selected from the visual or textual global token, corresponding to the Attention Pooling module and the Cross-Modality Attention Pooling module in Fig. 2(c). The visualization of attention maps from different attention heads are shown in Fig. 3. Fig. 3(b) shows that our Cross-Modality Attention Pooling module can align the global style words (like ``realistic style" and ``city") with the corresponding regions of image. And Fig. 3(c) illustrates the capability to capture the object words in the image such as ``alien, driving aircraft".

Finally, simply concatenate these two outputs and pass the result into a linear head to get the score. For the coupled MOS score commonly like AGIQA-1k\cite{zhang2023perceptual}, we just output one scalar. For decoupled scores like AGIQA-3k\cite{li2023agiqa}, our model predict two scores including the visual subjective quality and image-prompt alignment. 

\begin{figure}[!t]
\centering
\includegraphics[width=1\linewidth]{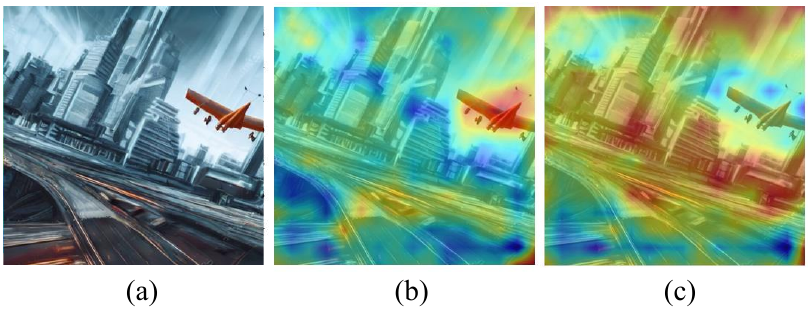}
\caption{Visualization of the attention maps within the Cross-Modality Attention Pooling module. From left to right: (a) input image; (b) highlights to the object-specific word ``aircraft"; (c) highlights to the scene-specific word ``city".}
\label{attention_map}
\end{figure}

%% file: experiment.tex
\section{Experiments}
\label{sec:experiment}

\subsection{Experimental Settings}

% \subsubsection{Incremental Pre-training}
% The data preparation of our incremental pretraining is performed on DiffusionDB\cite{wang2022diffusiondb}, a AGI database containing 14M Text-Image pairs generated by the Stable-Diffusion model. Initially, we use its subset, containing 2M AIGC image-text pairs, then we use CLIP to pre-calculate the cosine similarity of these image-text pairs and select samples above a certain threshold based on this. These pairs, serving as training data, are used to incrementally pretrain our image encoder. The similarity distribution is illustrated in Fig.\ref{sim_dist}.

% In this work, we set the threshold to 0.35 and retain 560K image-text pairs as our final pretraining data.

% \begin{figure}[htbp]
%     \centering
%     \includegraphics[scale=0.65]{figs/similarity_distribution.png}
%     \caption{The cosine similarity distribution of 2M image-prompt pairs from DiffusionDB subset.}
%     \label{sim_dist}
% \end{figure}

\subsubsection{Datasets}
Our quality evaluation experiments are conducted on two AGI subjective quality labeled databases, AGIQA-1k\cite{zhang2023perceptual} and AGIQA-3k\cite{li2023agiqa}. AGIQA-1k is the first subjective database containing 1,080 AGIs, which are generated by two T2I generation models \cite{rombach2022high} \textit{stable-inpainting-v1} and \textit{stable-diffusion-v2}. There are 180 prompts designed as input, and these prompts are merely simple combinations of high-frequency keywords extracted from image websites. In the subjective quality assessment process of AGIQA-1k, the perceptual quality of the generated images and the text-image correspondence are considered simultaneously to obtain MOS in the range of [1, 5].
AGIQA-3k is another AGI subjective database composed of 2,982 AGIs labeled with perceptual quality scores and text-image alignment scores separately. Compared to AGIQA-1k, more T2I generation models are employed to create this database. Besides, there are 300 prompts designed for generation models in order to cover a large number of real user inputs. The perceptual quality MOS and the text-image alignment MOS are both in range of [0, 5].

\begin{table*}[!t]
\centering
\caption{The performance results of perceptual models.}
\label{tab:perceptual}
\resizebox{1\linewidth}{!}{
    \setlength\tabcolsep{25pt}
\begin{tabular}{@{}cl|ccc|ccc@{}}
\toprule
\multicolumn{2}{c|}{\multirow{2}{*}{Methods}}    & \multicolumn{3}{c|}{AGIQA-1k} & \multicolumn{3}{c}{AGIQA-3k} \\ \cmidrule(l){3-8} 
\multicolumn{2}{c|}{}                & SRCC     & PLCC     & KRCC     & SRCC      & PLCC      & KRCC      \\ \midrule
\multirow{4}{*}{\makecell[c]{Handcrafted-\\-based}}  
                    & CEIQ          & 0.3069   &  0.2836  &  0.2097    & 0.3228    & 0.4166    &  0.2220      \\
                    & NIQE          & -0.5490   &  -0.5048  &  -0.3824    & 0.5623    & 0.5171    &  0.3876      \\
                    & DSIQA         & -0.3047   &  -0.0559  &  -0.2148    & 0.4955    & 0.5488    &  0.3403      \\
                    & SISBLIM       & -0.1309   &  -0.3575  &  -0.0889    & 0.5479    & 0.6477    &  0.3788      \\ \midrule
\multirow{2}{*}{SVR-based}
                    & GMLF          & 0.5575   &  0.6356  &  0.4052    & 0.6987    & 0.8181    &  0.5119      \\
                    & HIGRADE       & 0.4056   &  0.4425  &  0.2860    & 0.6171    & 0.7056    &  0.4410      \\ \midrule
\multirow{7}{*}{DL-based}
                    & DBCNN         & 0.7491  & 0.8211  &  0.5618    & 0.8207    & 0.8759    &  0.6336      \\ 
                    & CNNIQA        & 0.5800  &  0.7139 &  0.4095   & 0.7478    & 0.8469    &  0.5580      \\
                    & CLIPIQA       & 0.8227  & 0.8411  &  0.6399  & 0.8426 & 0.8053 & 0.6468 \\ 
                    & HyperNet      & 0.7803  &  0.8299  &  0.5943  & 0.8355    & 0.8903    &  0.6488   \\ 
                    & ResNet50    & 0.7136   & 0.7576   &  0.5254  & 0.6744    & 0.7365   & 0.4887   \\
                    & ResNet50*    & 0.7914   & 0.8404  & 0.6064   & 0.8306   &  0.8929  & 0.6465   \\
                    & Ours          & \textbf{0.8401}   & \textbf{0.8922}   & \textbf{0.6635}
                                     & \textbf{0.8634}   & \textbf{0.9116}   & \textbf{0.6844}   \\ \bottomrule
\end{tabular}}
\end{table*}

\subsubsection{Implementation Details}

The data preparation for our \textit{Image2Prompt} incremental pretraining is performed on one 2M subset of DiffusionDB\cite{wang2022diffusiondb}, an AGI database containing 14M Text-Image pairs generated by the Stable-Diffusion model. And 560K image-text pairs is selected according to the image-text cosine similarity calculated by CLIP above a certain threshold, which is 0.35 in this work. This strategy ensures that our incremental pretrain procedure is informed by high-quality, well-matched image-prompt pairs, thereby enhancing the efficacy of the model in capturing the interplay between the two modalities.

As for perception training, similar to the settings in \cite{zhang2023perceptual}, the two databases are both split randomly in an 80/20 ratio for training/testing while ensuring the image with the same object label falls into the same set. The partitioning and evaluation process is repeated 10 times for a fair comparison while considering the computational complexity, and the average result is reported as the final performance. The Adam optimization with a mini-batch of 40 and an initial learning rate of 1e-5 is adopted for training without decaying. All the images in both databases are kept in their original resolution $512 \times 512 \times 3$. Our model is trained for 100 epochs on a single NVIDIA GTX 3090 GPU.

The metrics adopted to evaluate our assessment model and benchmark models are Pearson Linear Correlation Coefficient (PLCC), Spearman Rank Correlation Coefficient (SRCC), and Kendall’s Rank Correlation Coefficient (KRCC), which measure the correlation between the predicted scores and corresponding subjective quality scores. And methods with PLCC, SRCC and KRCC closer to 1 are better.

\subsection{Experiment Results}
In terms of perception, our method is compared with 12 perceptual quality methods, including handcrafted-based metrics CEIQ\cite{yan2019no}, NIQE\cite{mittal2012making}, DSIQA\cite{narvekar2009no}, SISBLIM\cite{gu2014hybrid}
support vector regression-based metrics GMLF\cite{xue2014blind}, HIGRADE\cite{kundu2017large} and deep learning-based (DL) metrics DBCNN\cite{zhang_blind_2020}, CNNIQA\cite{kang2014convolutional}, HyperNet\cite{su2020blindly}, ResNet50 and ResNet50* (ImageNet pretrained) \cite{He_2016_CVPR}.
For alignment, we select four metrics CLIPScore\cite{hessel2021clipscore}, ImageReward\cite{xu2023imagereward}, HPS\cite{wu2023better} and StairReward\cite{li2023agiqa}.

Table \ref{tab:perceptual} lists the performance results of different perception models. Our method performs best on both databases, outperforming the second place method by 2.1\% and 2.4\% regarding the SRCC criterion respectively. As illustrating, handcrafted-based IQA methods show poor performance in evaluating AGIs, because the prior knowledge based on NSIs is not suitable for evaluating AGIs, which shows the domain gap between them. Deep learning-based methods show promising performance with an overall SRoCC larger than the other two type methods. Compared with the performance on AGIQA-3k database, the selected methods perform relatively poorly on AGIQA-1k database, especially the Handcrafted-based methods. And it is because the quality annotation of AGIQA-1k database considers both image perceptual quality and image-text correspondence. This phenonmenon also illustrates that unimodal models are difficult to evaluate the image-text correspondence of the input.

\begin{table}[!t]
\centering
\caption{The performance results of alignment models}
\label{tab:alignment}
\resizebox{1\linewidth}{!}{
    \setlength\tabcolsep{37pt}
\begin{tabular}{@{}r|ccc@{}}
\toprule
{\multirow{2}{*}{Method}}   & \multicolumn{3}{c}{Alignment} \\ \cmidrule(l){2-4} 
               & SRCC     & PLCC     & KRCC          \\ \midrule
CLIPScore      & 0.5972   & 0.6839   & 0.4591   \\
ImageReward    & 0.7298   & 0.7862   & 0.5390   \\
HPS            & 0.6349   & 0.7000   & 0.4580   \\
StairReward    & 0.7472   & 0.8529   & 0.5554   \\
Ours           & \textbf{0.7578}   & \textbf{0.8544}   & \textbf{0.5734} \\
\bottomrule
\end{tabular}}
\end{table}

\begin{table}[!t]
\centering
\caption{The ablation results on perceptual quality.}
\label{tab:ablation}
\resizebox{1\linewidth}{!}{
    \setlength\tabcolsep{2pt}
\begin{tabular}{@{}ccc|cccccc@{}}
\toprule
\multirow{2}{*}{Image2prompt} & \multirow{2}{*}{Integral prompt} & \multirow{2}{*}{\textit{[QA]} token} & \multicolumn{3}{c}{AGIQA-1k}                  & \multicolumn{3}{c}{AGIQA-3k} \\ \cmidrule(l){4-9} 
                              &                                  &                           & SRCC   & PLCC   & \multicolumn{1}{c|}{KRCC}   & SRCC     & PLCC    & KRCC    \\ \midrule
                              &                                  &                           & 0.8105 & 0.8595 & \multicolumn{1}{c|}{0.6173} & 0.8398   & 0.8978  & 0.6561  \\ \midrule
\checkmark                       &                                  &                           & 0.8180 & 0.8703 & \multicolumn{1}{c|}{0.6501} & 0.8491   & 0.9031  & 0.6690  \\ \midrule
                              & \checkmark                          &                           & 0.8317 & 0.8706 & \multicolumn{1}{c|}{0.6532} & 0.8442   & 0.9008  & 0.6612  \\ \midrule
\checkmark                       & \checkmark                          &                           & 0.8383 & 0.8782 & \multicolumn{1}{c|}{0.6603} & 0.8595   & 0.9084  & 0.6798  \\ \midrule
\checkmark                       & \checkmark                          & \checkmark                   & \textbf{0.8401} & \textbf{0.8922} & \multicolumn{1}{c|}{\textbf{0.6635}} & \textbf{0.8634}   & \textbf{0.9116}  & \textbf{0.6844}  \\ \bottomrule
\end{tabular}}
\end{table}

Table \ref{tab:alignment} lists the performance results of selected alignment models. The results show that our model performs best on AGIQA-3k database, and it outperforms the second place method 1.4\% regarding SRCC metric. It is worth mentioning that with the \textit{Image2Prompt} and image-prompt fusion module, our proposed method can predict the alignment of prompt in different length.

\subsection{Ablation Study}
To verify the effectiveness of our proposed method, we conduct ablation study on the two databases. The perceptual results of the ablation studies on the two databases are illustrated in Tab \ref{tab:ablation}. The first line shows the performance of our baseline, which is ResNet50 initialized by CLIP\cite{radford2021learning} image encoder.

% \usepackage{booktabs}
% \usepackage{multirow}

\noindent\textbf{Impact of \textit{Image2Prompt}}. The \textit{Image2Prompt} is proposed to bridge the AGI-style visual and textual modalities. 
Without \textit{Image2Prompt}, the performance of baseline model decreases 0.9\% on AGIQA-1k and 1.1\% on AGIQA-3k. 
Compared to the performance of the model with the \textit{Image2Prompt} and integrating with textual prompt, that of the model only integrating with the textual prompt drops 0.8\% on AGIQA-1k and 1.8\% on AGIQA-3k. 
These results show that our \textit{Image2Prompt} can enhance the multimodal understanding.

\noindent\textbf{Impact of Integral Prompt}. The integration of textual prompts can help assess the correspondence between the image and its textual prompt. The absence of integral prompt leads to two decreases on both two databases, which are 2.4\% and 0.5\% respectively. The performance changes show the necessity of introducing text modality for AGIQA.

\noindent\textbf{Impact of \textit{[QA]} Token}. We introduce the \textit{[QA]} token to make the network concentrate on vocabulary related to image quality during the extraction of prompt embedding. The absence of this module makes the performance drop by 0.2\% on AGIQA-1k and 0.5\% on AGIQA-3k. In a consequence, a specific \textit{[QA]} token in the prompt textual domain can make model focus on quality-relevant aspects of both image and its corresponding prompt.

%% file: conclusion.tex
\section{Conclusion}
\label{sec:conclusion}

We propose a multimodal framework named IP-IQA for AGIQA in this paper. We apply an incremental pretraining method, Image2Prompt, for better understanding of AGIs and their corresponding prompts. In order for the integral prompt, we introduce a modular image-prompt fusion component based on cross-attention, along with the novel \textit{[QA]} token. The visualization in Fig. 3 and our experiments illustrate their effectiveness and how they work. Our IP-IQA achieves the state-of-the-art of AGIQA-1k and AGIQA-3k, working well on not only the image quality but also the alignment of image and prompt. We think that IP-IQA aligns with the inherent multimodal nature of AGIs from birth and really hope that IP-IQA is able to be a good multimodal reference for further research purpose on AGIQA community. The limitation of IP-IQA lies in its lack of consideration for the deeper relationships between images and their corresponding prompts. We will serve it as our direction for future improvements.